\documentclass[runningheads]{llncs}
\usepackage[T1]{fontenc}
\usepackage{graphicx}
\usepackage{color}
\usepackage{subcaption}
\usepackage[english]{babel}
\usepackage[utf8]{inputenc}
\usepackage{algorithm}
\usepackage{arevmath}     
\usepackage[noend]{algpseudocode}
\usepackage{amsmath}
\usepackage{booktabs,dcolumn}
\usepackage{hyperref}
\begin{document}
\title{Recommender Engine Driven Client Selection in Federated Brain Tumor Segmentation}
\titlerunning{Recommender Engine for Client Selection in Federated Learning}
%
\author{Muhammad Irfan Khan\inst{1} \and
Elina Kontio\inst{1} \and
Suleiman A. Khan\inst{1} \and
Mojtaba Jafaritadi\inst{1}}
\authorrunning{Khan et al., 2024}
%
\institute{Turku University of Applied Sciences, Turku 20520, Finland 
\\
\email{irfan.khan, elina.kontio, suleiman.alikhan, mojtaba.jafaritadi@turkuamk.fi}
} 
\maketitle              
\begin{abstract}
This study presents a robust and efficient client selection protocol designed to optimize the Federated Learning (FL) process for the Federated Tumor Segmentation Challenge (FeTS 2024). In the evolving landscape of FL, the judicious selection of collaborators emerges as a critical determinant for the success and efficiency of collective learning endeavors, particularly in domains requiring high precision. This work introduces a recommender engine framework based on non-negative matrix factorization (NNMF) and a hybrid aggregation approach that blends content-based and collaborative filtering. This method intelligently analyzes historical performance, expertise, and other relevant metrics to identify the most suitable collaborators. This approach not only addresses the cold start problem where new or inactive collaborators pose selection challenges due to limited data but also significantly improves the precision and efficiency of the FL process. Additionally, we propose harmonic similarity weight aggregation (HSimAgg) for adaptive aggregation of model parameters. We utilized a dataset comprising 1,251 multi-parametric magnetic resonance imaging (mpMRI) scans from individuals diagnosed with glioblastoma (GBM) for training purposes and an additional 219 mpMRI scans for external evaluations. Our federated tumor segmentation approach achieved dice scores of 0.7298, 0.7424, and 0.8218 for enhancing tumor (ET), tumor core (TC), and whole tumor (WT) segmentation tasks respectively on the external validation set. In conclusion, this research demonstrates that selecting collaborators with expertise aligned to specific tasks, like brain tumor segmentation, improves the effectiveness of FL networks. 


\end{abstract}
\section{Introduction}
\label{sec:intro}
Federated learning (FL) is a collaborative setting that harnesses the collective power of distributed computation and data sources in solving a machine learning problem \cite{Kairouz2019}. It facilitates the gathering of decentralized knowledge and insights from multiple entities or gamut of data repositories while maintaining data privacy and security. In essence, FL involves a continuous flow of knowledge gained from data of multiple institutions without pooling their data together. This collaborative model training strategy includes decentralized data silos, a central server, model initialization and deployment, local model training, model update exchange, and iterative cycles of parameter aggregations \cite{zhao2018federated,Kairouz2019,mcmahan2017communication}.  By decentralizing the learning process, FL enables collaboration among multiple data sources while restricting direct data sharing. This approach is particularly useful for sensitive and distributed data.

Selection of collaborators in FL is crucial as it allows maximizing the learning process and guarantee the effectiveness of the collaborative model. However, collaborator selection in FL faces challenges stemming from data heterogeneity (IID vs. Non-IID) across clients \cite{zhao2018federated}, varying system resources \cite{nishio2019client,li2020federated}, scalability issues, privacy concerns \cite{Kairouz2019}, incentive mechanisms for participation \cite{zhan2020learning}, and dynamic client availability \cite{KhanSimAgg}. These challenges arise due to the decentralized nature of FL, where multiple clients collectively train a shared model while keeping their local data private. Addressing these challenges is crucial for efficient and effective model training, and researchers have proposed various techniques, including clustering-based approaches \cite{sattler2020clustered}, resource-aware selection \cite{sattler2019robust}, privacy-preserving selection \cite{truex2019hybrid}, and incentive-based selection mechanisms \cite{zhu2024client,zhan2020learning}. However, developing a unified and robust collaborator selection strategy that comprehensively tackles all these challenges remains an active area of research. Recommender systems for collaborator selection have been previously introduced to deal with these challenges and enhance the model's accuracy and efficiency \cite{ji2024emerging}.

In this paper, we propose a recommender engine system for the federated brain lesion segmentation task which, unlike the conventional methods like batch-wise selection methods of collaborator selection, analyzes retrospective performance data, expertise, and other relevant metrics. We consider a non-negative matrix factorization (NNMF) strategy to enhance the recommender system in the context of collaborator selection in FL. By decomposing complex datasets into simpler, interpretable matrices while preserving the inherent non-negativity of data, NNMF effectively uncovers latent features by factorizing into low-rank approximations that represent underlying patterns in collaborator performance and interactions. In the FL environment, where selecting the most suitable collaborators is crucial for tasks like brain tumor segmentation, NNMF can analyze historical performance, skills, and other relevant metrics to identify these latent attributes. This advanced analytical capability enables a more nuanced and informed selection process, ensuring that each collaborator's contribution aligns optimally with the specific requirements of the learning task. Our results show that the integration of NNMF-based recommender engines in FL not only enhances the precision of collaborator selection but also significantly boosts the overall efficiency and efficacy of the federated model, leading to more accurate and reliable outcomes.

Our main contributions in this paper are 1) the establishment of NNMF strategy for a recommender engine system that efficiently and dynamically selects the collaborators; 2) the implementation of a new adaptive weight aggregation algorithm that can be applied to this setting; and 3) an extensive evaluation of the proposed weight aggregation approach. 

This paper is organized as follows: in Section~\ref{sec:method}, we describe the methodologies including our two FL weight aggregation strategies by our experiment setting. In Section~\ref{sec:results} we describe FL experiments and evaluate the performance of the proposed methods quantitatively and in Section~\ref{sec:discussion} we discuss about the presented work, potentials and limitations, and describe our future direction in FL. Finally, Section~\ref{sec:conclusion} concludes this work.

\section{Methods}
\label{sec:method}
\subsection{Data}
Our research utilized training data provided by the FeTS Challenge 2022 \cite{spyridon_bakas_2022_6362409}. This dataset encompassed a total of 1251 multi-parametric magnetic resonance imaging (mpMRI) scans acquired from individuals diagnosed with glioblastoma (GBM), an aggressive form of brain cancer. The mpMRI scans consisted of four modalities: native (T1-weighted), post-contrast T1-weighted (T1Gd), T2-weighted (T2), and T2 Fluid-Attenuated Inversion Recovery (FLAIR) volumes. It's important to note that the FeTS 2022 data represents a subset of GBM cases drawn from the Brain Tumor Segmentation (BraTS) continuous challenge ~\cite{baid2021rsna,bakas2017segmentationGBM,bakas2017segmentationLGG,bakas2017advancing,menze2014multimodal}. The BraTS challenge focuses on identifying state-of-the-art algorithms for segmenting brain diffuse gliomas and their sub-regions.

The mpMRI data originated from various universities and underwent preprocessing steps to ensure consistency. These steps included rigid registration for spatial alignment, brain extraction to remove non-brain tissue, intensity normalization for voxel value harmonization, resolution resampling for standardized image dimensions, skull stripping for background removal.

\subsection{Federated Learning Framework}

To train a brain tumor segmentation model while preserving data privacy, we employed Intel's Federated Learning (OpenFL) framework \cite{reina2021openfl}. This framework facilitated collaborative learning without requiring participants to share their raw data. Additionally, we utilized a 33 million parameter, 95 layered U-Net convolutional neural network architecture provided by the FeTS 2022 competition \cite{spyridon_bakas_2022_6362409}. The FL process involved a designated central server coordinating the training process. In each training round, a subset of collaborators (20\% of participants) contributed their locally updated model parameters to the central server. The central server then aggregated these updates and broadcasted the improved global model back to the participants for the next training iteration. This collaborative learning methodology ensured privacy protection while enabling the development of a robust brain tumor segmentation model.

\subsubsection{Recommender Engine Collaborator Selection}
\label{m1}
{In our previous work on the FeTS 2021 challenge, we introduced an batch-wise collaborator selection policy \cite{KhanSimAgg}. This method selects a subset of the total available collaborators (e.g., 20\%) in each round. To accommodate system heterogeneity, where collaborators contribute in a non-deterministic manner, we simulated random selection of collaborators and implemented a sliding window over the randomized collaborator index in each round.

Building on this, we now present a novel recommender engine system for collaborator selection. This system selects 20\% of the collaborators in each federation round. However, it introduces a distinctive approach: in one round, the top 20\% of collaborator recommendations are chosen, while in the next round, the bottom 20\% are selected. This novel system enhances efficiency and dynamism in client selection, making federated weight aggregation less susceptible to outliers or extreme values. 
}



The proposed recommender engine workflow deploys NNMF on historical performance data to evaluate past contributions and unveil underlying patterns in collaborators' efficiency and reliability. It considers quantitative performance metrics and behavioral factors (participation frequency and total contribution time), ensuring a comprehensive selection process. 

\begin{figure*}[!th]
\centering\includegraphics[width=0.9 \textwidth]{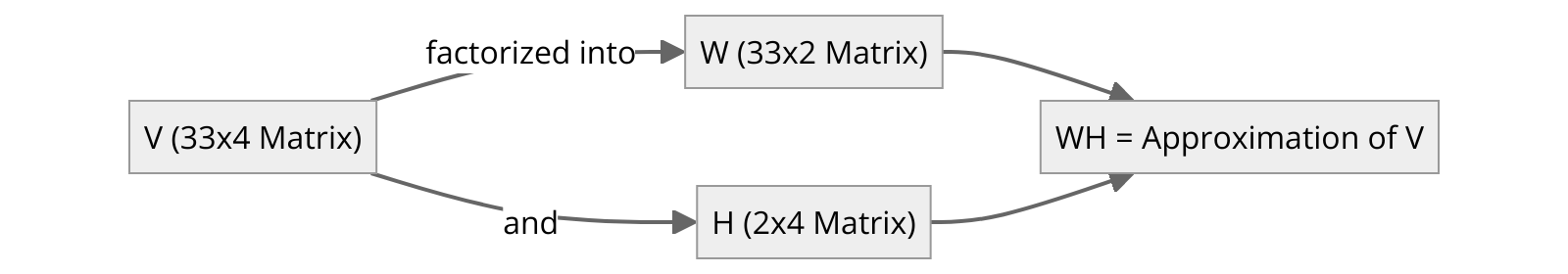}
\caption{Non-negative matrix factorization (NNMF) strategy.}
\label{fig:NNMF}
\end{figure*}

The collaborator selection process in the proposed recommender engine system involves filtering the database records from the previous FL round to gather aggregate validation metrics, such as the Dice coefficient and loss, for evaluating each collaborator's performance. Additionally, the frequency of selection and total contribution time across previous rounds are retrieved to identify consistently active participants. The decision-making metrics include performance metrics (Dice coefficient and Loss), frequency of participation, and total contribution time. 
 { Dice score reflects segmentation accuracy, while Loss ensures the selection process accounts for model convergence consistency. This combination adds robustness to the collaborator selection process, balancing accuracy and learning consistency, especially when segmentation accuracy and model stability do not directly correlate.} For our experiments on partition 2 of the FeTS22 data where there are 33 collaborators, a matrix of dimensions 33 $\times$ 4 is maintained, containing these metrics. The data is then normalized to ensure fair comparison across different scales. The NNMF with 2 components is employed to decompose the normalized data into matrices representing latent features. This step is crucial for identifying underlying patterns in the collaborators' performance and contribution behaviors, resulting in matrices of dimensions 33 $\times$ 2 and 2 $\times$ 4 (see Fig. \ref{fig:NNMF}). 

{For collaborator selection 
the algorithm balances exploration and exploitation through odd and even rounds. In odd rounds, it prioritizes exploring newer or less frequently selected collaborators by giving more weight to those with lower contributions in previous rounds, based on their NNMF scores. In even rounds, it focuses on exploiting high-performing collaborators to reinforce learning with consistent contributions. This balance is achieved by using the first latent variable from the NNMF decomposition, which represents the most important performance pattern, ensuring fair selection of both new and high-performing participants. The NNMF dynamically ranks collaborators based on their latent features, optimizing both exploration and exploitation phases.}

As a single point of fail-safe fallback mechanism, a random selection of collaborators is integrated. If the NNMF-based selection process does not yield a result (e.g., due to insufficient data), the function falls back to this random selection. This ensures that the FL process can continue even in the absence of detailed historical performance data. The adaptive nature of this workflow dynamically adjusts its selection criteria based on the current FL round, striking a balance between exploring new collaborators and exploiting known high-performers. This allows a collaborative environment where optimizing collective output is crucial, such as in federated segmentation tasks.


Finally, a list of collaborators for the current round is generated, which can be used to coordinate the subsequent steps in the FL process. This approach allows for a dynamic and data-driven selection of collaborators, aiming to optimize the learning process by leveraging historical performance data and machine learning techniques to predict which collaborators are likely to offer the most valuable contributions in future rounds.





 \subsubsection{Harmonic Similarity Weighted Aggregation (HSimAgg) }
This work introduces the Harmonic Similarity Weighted Aggregation (HSimAgg) algorithm, an advancement of the SimAgg algorithm tailored for brain tumor segmentation. HSimAgg, in conjunction with a dynamic collaborator selection strategy, leverages the harmonic mean for the final aggregation of model parameters. This choice enables robust handling of outliers or extreme values within the dataset.

A critical challenge in federated learning with non-IID data arises from the potential divergence of model parameters contributed by collaborators. To address this issue, HSimAgg employs a weighted aggregation approach at the server. Collaborators are assigned weights based on their harmonic similarity to the unweighted average model. This intuitive yet effective mechanism facilitates the creation of a master model that effectively captures the consensus of the majority of collaborators during each iteration. The detailed workings of the aggregation algorithm are presented in Algorithm~\ref{simagg_algo}.


\begin{algorithm}
\caption{HSimAgg aggregation algorithm}
\label{simagg_algo}
\begin{algorithmic}[1]
\Procedure{Weight Aggregation}{$C^r$, $p_{C^r}$}

    \State $\epsilon$ $\leftarrow$ $1e-5$ \Comment{$C^r$ = set of collaborators (at round $r$)}
    
    \State  $\hat{p}$ = average($p_{C^r}$) using \textbf{Eq.~\ref{eqn_pavg}} \Comment{$p_{C^r}$ = parameters of the collaborators in $C^r$}
    
    \For{$c$ in $C^r$} 
        \State Compute similarity weights $u_c$ using \textbf{Eqs.~\ref{eqn_sim}} and \textbf{\ref{eqn_sim_weights}}
        \State Compute sample weights $v_c$ using \textbf{Eq.~\ref{eqn_sample_weights}}
    \EndFor 
    
    \For{$c$ in $C^r$} 
        \State Compute aggregation weights $w_c$ using \textbf{Eq.~\ref{eqn_fsim_weights}}
    \EndFor  
    
    
    \State Compute master model parameters $p^m$ using \textbf{Eq.~\ref{eqn_params}}
    
    \State \textbf{return} $p^m$
\EndProcedure
\end{algorithmic}
\end{algorithm}

During round $r$, the server receives the parameters $p_{C^r}$ contributed by the collaborating entities $C^r$. Subsequently, the server computes the average of these parameters as follows

\begin{equation}
\hat{p} = \frac{1}{\lvert C^r \lvert}\Sigma_{i\in C^r}{p_i}.
\label{eqn_pavg}
\end{equation}

Next, we proceed to determine the inverse distance (similarity) of each collaborator $c$ within $C^r$ from the calculated average

\begin{equation}
sim_c = \frac{\Sigma_{i \in C^r}{\lvert p_i - \hat{p} \rvert}}{\lvert p_c - \hat{p} \rvert + \epsilon},
\label{eqn_sim}
\end{equation}
where $\epsilon = 1e-5$ (small positive constant). We standardize the distances to derive the "similarity weights" in the subsequent manner

\begin{equation}
u_c = \frac{sim_c}{\Sigma_{i\in C^r}{sim_i}}.
\label{eqn_sim_weights}
\end{equation}

HSimAgg assigns higher weights to collaborators whose model parameters exhibit greater harmonic similarity to the unweighted average model. Conversely, collaborators with significant deviations receive lower weights. This weighting strategy effectively reduces the influence of outliers or data points with substantial divergence, minimizing their impact on the final aggregated model.


To accommodate the varying influence of distinct sample sizes across each collaborator $c$  within $C^r$, we employ "sample size weights" that prioritize collaborators with a greater number of samples.

\begin{equation}
v_c = \frac{N_c}{\Sigma_{i\in C^r}{N_i}},
\label{eqn_sample_weights}
\end{equation}
where $N_c$ is the number of examples at collaborator $c$. 

Using the weights obtained using Eqs.~\ref{eqn_sim_weights} and~\ref{eqn_sample_weights}, the {\em aggregation weights} are computed as:

\begin{equation}
w_c = \frac{u_c+v_c}{\Sigma_{i\in C^r}{(u_i+v_i)}},
\label{eqn_fsim_weights}
\end{equation}

Ultimately, the aggregation of parameters is done through the harmonic mean of the aggregation weights. The harmonic mean, a statistical average that accommodates situations with reciprocal relationships, is harnessed to consolidate the parameters effectively. This choice of aggregation method enables the model to consider the contributions of collaborators more holistically, especially in scenarios where divergent or extreme values might be present in the dataset. The harmonic mean operates by calculating the reciprocal of the arithmetic mean of the reciprocals of the values, and thus it is suitable for cases where proportionality and balance are key factors.

\begin{equation}
p^m = \frac{1}{\sum_{i\in C^r}{\frac{w_i}{p_i}}} \cdot \Sigma_{i\in C^r}{(w_i \cdot p_i)}.
\label{eqn_params}
\end{equation}

In the following rounds of federation, the normalized aggregated parameters  $p^m$ are extended to the subsequent cohorts of collaborators.

\section{Experiments}
\label{sec:results}

\subsection{Setup}
The experimental framework is designed to optimize the FL process by streamlining model aggregation, client selection, round-specific training, and communication strategies. The primary goal is to achieve optimal collaborator selection using NNMF strategy, combined with efficient aggregation of model updates from specific contributors. For this study, a comprehensive dataset of 1251 patients from multiple institutions was used for training, while an additional set of 219 patients served as the validation cohort. The collaborative effort involved 33 participants responsible for partitioning the dataset.

The experiment utilized a pre-configured 3D U-shaped neural network, and the semantic segmentation tasks of the complete tumor, tumor core, and enhancing tumor were executed using Intel's OpenFL platform. The evaluation metrics included Binary DICE similarity and Hausdorff (95\% percentile) distance, providing a robust assessment of the effectiveness of the aggregation rounds, as described in~\cite{pati2021federated}. This experimental setup serves as a solid foundation for evaluating the performance improvements achieved through the proposed recommender engine-based collaborator selection and model aggregation strategies within the FL paradigm. The hyperparameters used are shown in Table~\ref{turns_Hyperparameters}. 

\begin{table}[H]
\centering
\caption{Hyperparameters used in aggregation algorithms.}
\label{turns_Hyperparameters}
\resizebox{0.5\columnwidth}{!}{%
\begin{tabular}{ll}
\hline
Hyperparameter       & HSimAgg \\  \hline
Learning rate        & 5e-5    \\
Epochs per round     & 1.0                                   \\
Communication rounds & 20                                  \\ \hline
\end{tabular}%
}
\end{table}

\subsection{Results}
In this section, a concise overview is presented for collaborator selection via reinforcement approaches using modified similarity weighted aggregation (HSimAgg). The findings underscore that the proposed techniques demonstrate rapid convergence and consistent stability as the learning process advances, encompassing all evaluated criteria.

\subsubsection{Model training and performance using internal validation data.}
\begin{figure}[htbp]
    \centering
    \begin{subfigure}[b]{0.45\textwidth}
        \includegraphics[width=\textwidth]{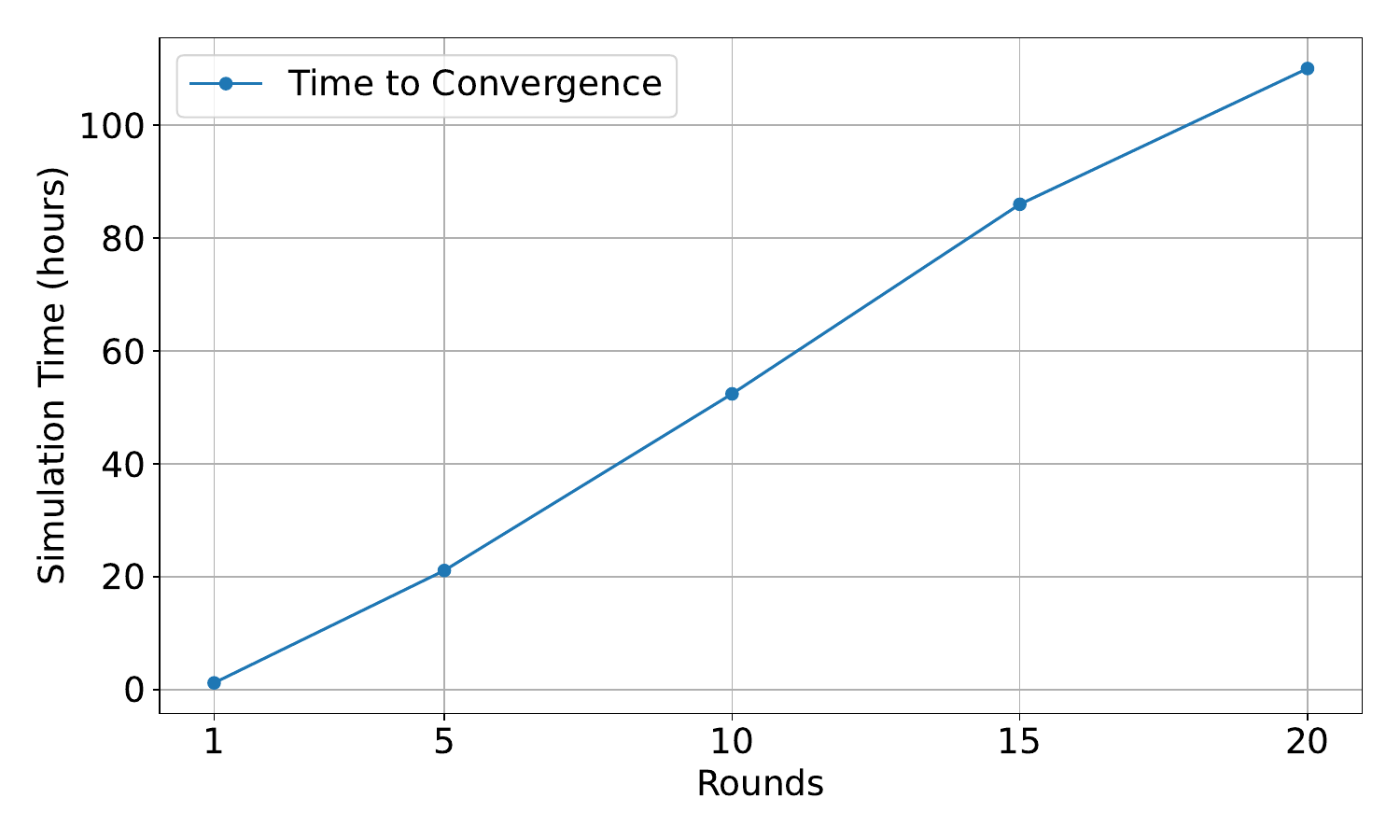}
        \caption{Simulation Time}
        \label{fig:sub1}
    \end{subfigure}
    \hfill
    \begin{subfigure}[b]{0.45\textwidth}
        \includegraphics[width=\textwidth]{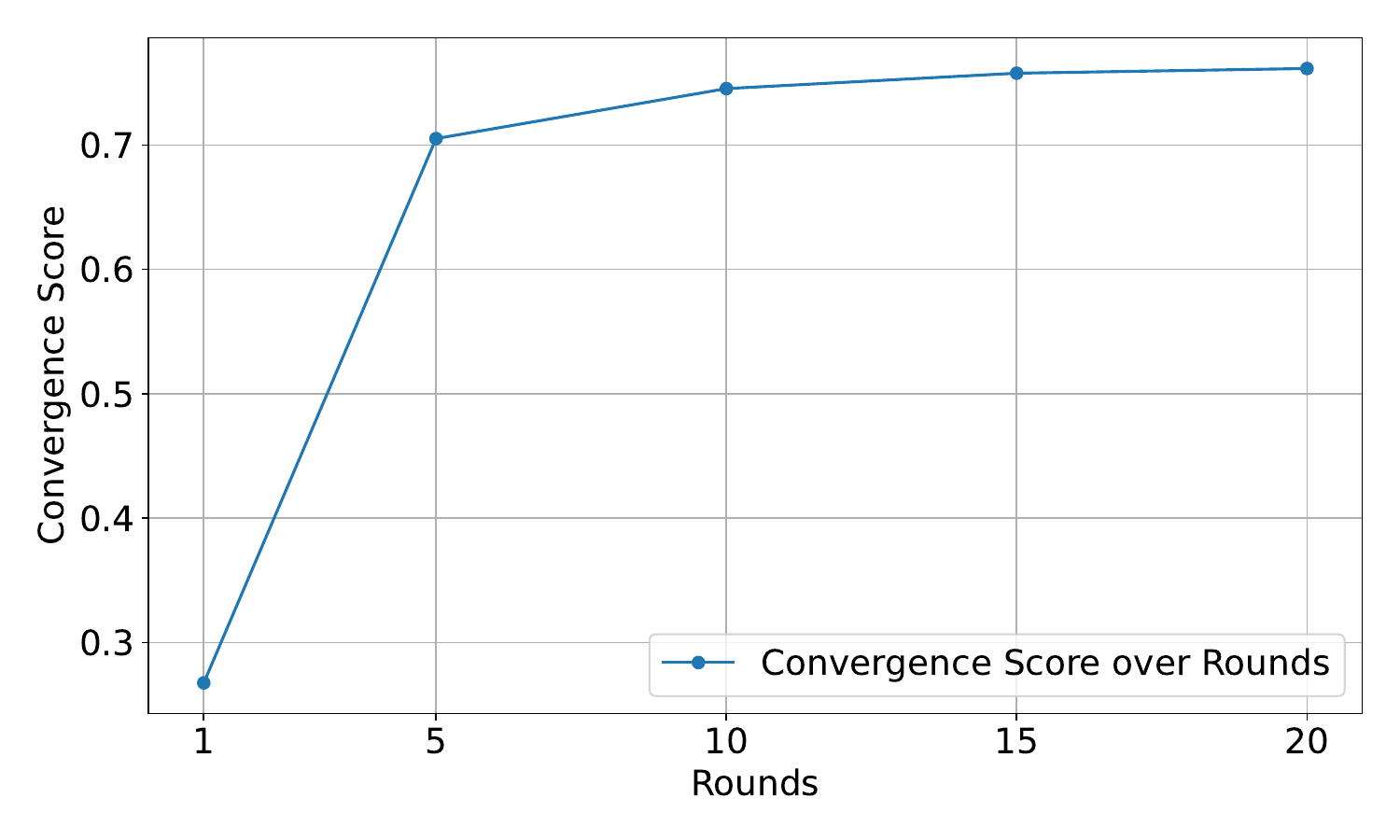}
        \caption{Projected Convergence Score}
        \label{fig:sub2}
    \end{subfigure}
    \\
    \begin{subfigure}[b]{0.45\textwidth}
        \includegraphics[width=\textwidth]{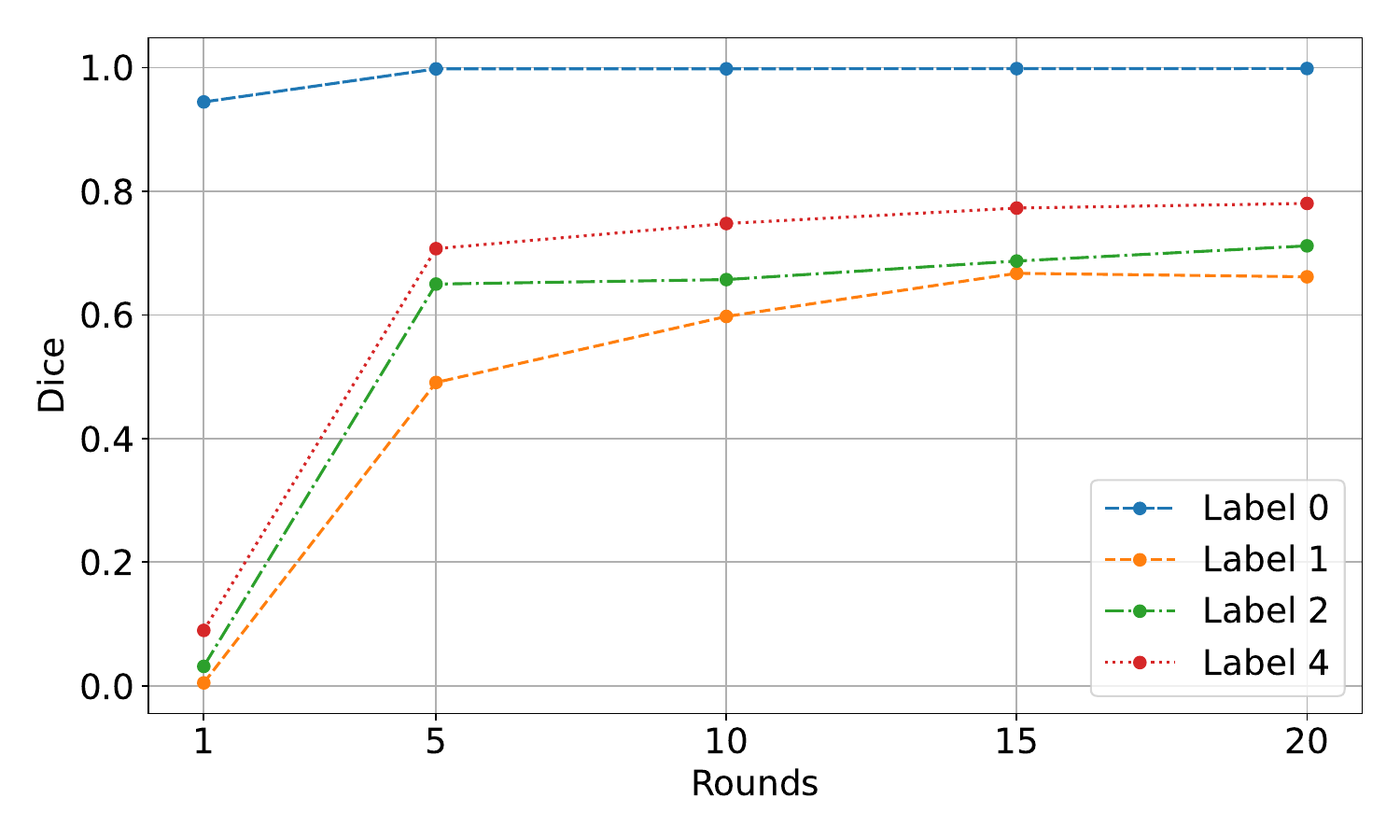}
        \caption{Dice similarity score}
        \label{fig:sub3}
    \end{subfigure}
    \hfill
    
    \caption{Performance metrics for model training of HSimAgg. The horizontal axis refers to the number of rounds and the vertical axis to the performance metrics.}
    \label{fig:foursubfigures}
\end{figure}

Figure~\ref{fig:foursubfigures} illustrates the model training performance on internal validation data observed over 20 rounds of federated model training calculating simulated time, convergence score and DICE scores. Training took 20:17:57 wall-clock time and consumed 306.87 GB memory using 3.96 KWh energy on GPU. The labels 0, 1, 2, and 4 are employed to characterize distinct classes concerning brain tumor segmentation. Specifically, label 0 denotes normal brain tissue, label 1 signifies the entirety of the tumor along with its surrounding swelling area, label 2 corresponds to the tumor core, and label 4 designates the enhancing region of the tumor.

\subsubsection{Model performance using external validation data.}
Our approach's performance is evaluated using 219 instances of external validation data. Table~\ref{tab:lesion} describes the model evaluation over the lesions. These metrics, lesion-wise dice score and lesion-wise Hausdorff distance, focus on how well a model performs on individual areas of abnormality (lesions) within an image, rather than the entire image itself. This allows for a more nuanced assessment by identifying how good the model is at detecting and outlining specific lesions, and avoids favoring models that only handle large abnormalities. Table~\ref{tab:brain} summarizes the model performance over the entire brain region. 

\begin{table}[th]
\centering
\caption{Lesion-wise performance evaluation on validation data.}
\label{tab:lesion}
\resizebox{0.5\columnwidth}{!}{%
\begin{tabular}{lc}
\hline
Metrics & HSimAgg  \\ \hline
Dice ET        & 0.4261409840   \\
Dice TC        & 0.4363372918   \\
Dice WT        & 0.1504901728   \\
Hausdorff (95\%) ET & 181.6372681  \\
Hausdorff (95\%) TC & 171.6629477  \\
Hausdorff (95\%) WT & 310.6808353  \\\hline
\end{tabular}%
}
\end{table}

\begin{table}[th]
\centering
\caption{Total brain performance evaluation on the external validation data.}
\label{tab:brain}
\resizebox{0.5\columnwidth}{!}{%
\begin{tabular}{lc}
\hline
Metrics & HSimAgg  \\ \hline
Dice ET        & 0.7297734504    \\
Dice TC        & 0.7424402893   \\
Dice WT        & 0.8217595748   \\
Hausdorff (95\%) ET & 33.80939079  \\
Hausdorff (95\%) TC & 26.08140704  \\
Hausdorff (95\%) WT & 35.57456896  \\
Sensitivity ET & 0.7006077605  \\
Sensitivity TC & 0.7096288676  \\
Sensitivity WT & 0.8440590033  \\
Specificity ET &  0.9997994167  \\
Specificity TC & 0.9997933843  \\
Specificity WT & 0.9986844926  \\ \hline
\end{tabular}%
}
\end{table}

\section{Discussion}
\label{sec:discussion}

The integration of a sophisticated recommender engine utilizing NNMF within a FL environment, as demonstrated in our study, marks an advancement in the field of collaborative learning, particularly in the context of medical image segmentation. The experimental results, as outlined in our dataset, underscore the engine's effectiveness in optimizing collaborator selection, thereby enhancing the model's accuracy and efficiency. By addressing the cold start problem through a hybrid approach that incorporates both content-based and collaborative filtering techniques, our framework showed a notable improvement in the selection process, even in scenarios with limited historical performance data.

The experiments highlighted a marked increase in the accuracy and efficiency of the FL model specially in a dynamic environment where data distribution within a collaborator is subject to change and fresh collaborators can join the federation network, with the recommender system successfully identifying collaborators whose expertise and historical performance align closely with the specific demands of the task. This alignment was crucial for tasks requiring high precision, resulting in improved outcomes for complex medical image analyses. Furthermore, the system's capability to dynamically adapt to new and inactive collaborators ensures the robustness and future scalability of the FL network. Another interesting problem is to handle the straggler collaborator, a potential solution is to use embarrassingly parallel method to parallelize processing to speed up the performance evaluation in straggler collaborator. Utilizing structured sparcity or quantum computing to make efficient communication protocols in federated learning is an interesting area of research as well.

In light of these promising findings, our study not only validates the efficacy of employing advanced recommender systems in FL environments but also opens avenues for further research into enhancing collaborative models through intelligent selection mechanisms. Future work could explore the integration of additional predictive metrics and real-world testing across diverse FL applications, aiming to solidify the foundation laid by this initial exploration and expand the potential of FL in critical, data-sensitive fields. Additionally, mitigating potential biases in historical data used by the recommender engine can further enhance its robustness. As FL continues to develop, our framework provides a strong foundation for efficient and targeted collaborative learning. This has the potential to revolutionize how we approach distributed machine learning tasks.

\section{Conclusion}
\label{sec:conclusion}

Our investigation into the application of a NNMF-based recommender engine for the strategic selection of collaborators in FL environments, specifically for medical image segmentation tasks, yields compelling outcomes. 
Our findings affirm the recommender engine's utility in optimizing collaborator selection, facilitating a more targeted and efficient allocation of resources in FL projects. By intelligently analyzing historical performance and expertise, the engine not only addresses the cold start problem but also ensures that each collaborator's contribution is optimally aligned with the specific needs of the task at hand. This strategic approach to collaborator selection is instrumental in advancing the precision and effectiveness of FL networks, especially in domains where accuracy is paramount.

\bibliographystyle{splncs04}
\bibliography{references.bib}
\end{document}